\documentclass[sigconf]{acmart}
\usepackage{amsmath}
\usepackage{enumitem}
\usepackage{multirow}
\usepackage{makecell}
\usepackage{stfloats}
\usepackage{colortbl}
\usepackage{balance}
\usepackage{bbm}

\AtBeginDocument{%
  \providecommand\BibTeX{{%
    \normalfont B\kern-0.5em{\scshape i\kern-0.25em b}\kern-0.8em\TeX}}}

\copyrightyear{2024}

\acmYear{2024}

\setcopyright{acmlicensed}

\acmConference[KDD '24] {Proceedings of the 30th ACM SIGKDD Conference on Knowledge Discovery and Data Mining }{August 25--29, 2024}{Barcelona, Spain.}

\acmBooktitle{Proceedings of the 30th ACM SIGKDD Conference on Knowledge Discovery and Data Mining (KDD '24), August    25--29, 2024, Barcelona, Spain}

\acmDOI{10.1145/3637528.3671838}

\acmISBN{979-8-4007-0490-1/24/08}

\begin{document}

\title{Unveiling Global Interactive Patterns across Graphs: Towards Interpretable Graph Neural Networks}

\author{Yuwen Wang}
\affiliation{%
  \institution{School of Software Technology, Zhejiang University}
  \institution{State Key Laboratory of Blockchain and Security, Zhejiang University}
  \institution{Hangzhou High-Tech Zone (Binjiang) Institute of Blockchain and Data Security}  
  \city{Hangzhou}
  \country{China}
}
\email{yuwenwang@zju.edu.cn}

\author{Shunyu Liu}
\authornote{Shunyu Liu is the corresponding Author.}
\affiliation{%
  \institution{State Key Laboratory of Blockchain and Security, Zhejiang University}
  \institution{Hangzhou High-Tech Zone (Binjiang) Institute of Blockchain and Data Security}  
  \city{Hangzhou}
  \country{China}
}
\email{liushunyu@zju.edu.cn}

\author{Tongya Zheng}
\affiliation{%
  \institution{Big Graph Center, School of Computer and Computing Science, Hangzhou City University}
  \institution{College of Computer Science and Technology, Zhejiang University}  
  \city{Hangzhou}
  \country{China}
}
\email{tyzheng@zju.edu.cn}

\author{Kaixuan Chen}
\author{Mingli Song}
\affiliation{%
  \institution{State Key Laboratory of Blockchain and Security, Zhejiang University}
  \institution{Hangzhou High-Tech Zone (Binjiang) Institute of Blockchain and Data Security}  
  \city{Hangzhou}
  \country{China}
}
\email{{chenkx, brooksong}@zju.edu.cn}

\renewcommand{\shortauthors}{Yuwen Wang, Shunyu Liu, Tongya Zheng, Kaixuan Chen, \& Mingli Song}

\begin{abstract}
{
Graph Neural Networks (GNNs) have emerged as a prominent framework for graph mining, leading to significant advances across various domains.
Stemmed from the node-wise representations of GNNs, existing explanation studies have embraced the subgraph-specific viewpoint that attributes the decision results to the salient features and local structures of nodes.
However, graph-level tasks necessitate long-range dependencies and global interactions for advanced GNNs, deviating significantly from subgraph-specific explanations.
To bridge this gap, this paper proposes a novel intrinsically interpretable scheme for graph classification, termed as Global Interactive Pattern (GIP) learning, which introduces learnable global interactive patterns to explicitly interpret decisions.
GIP first tackles the complexity of interpretation by clustering numerous nodes using a constrained graph clustering module.
Then, it matches the coarsened global interactive instance with a batch of self-interpretable graph prototypes, thereby facilitating a transparent graph-level reasoning process.
Extensive experiments conducted on both synthetic and real-world benchmarks demonstrate that the proposed GIP yields significantly superior interpretability and competitive performance to~the state-of-the-art counterparts. Our code will be made publicly available\footnote{
The code is available at \url{https://github.com/Wangyuwen0627/GIP-Framework.git}.}.
}
\end{abstract}




\maketitle
\section{Introduction}
Graphs, serving as data structures capable of naturally modeling intricate relationships between entities, have pervasive applications in real-world scenarios, such as transportation networks~\cite{chen2019gated,zhang2021traffic}, social networks~\cite{ying2018graph,liu2022federated}, power system~\cite{liu2023transmission, chen2022distribution2, chen2024powerformer}, and biological molecules~\cite{hao2020asgn,rong2020self}.
In recent years, to effectively uncover potential information in graphs for applications, graph neural networks~(GNNs)~\cite{hamilton2017inductive, kipf2016semi, velivckovic2018graph, chen2023improving}
have emerged as a prominent paradigm and made remarkable achievements.
Following a message-passing mechanism, GNNs aggregate the information from the local neighbors of each node to obtain node-wise representations, bolstering the development in various downstream tasks including node classification~\cite{zhao2021graphsmote,liu2022calibration, chen2023improving, wang2023adversarial} and graph classification~\cite{hamilton2017inductive, kipf2016semi, velivckovic2018graph, chen2022distribution}.

Despite the remarkable effectiveness of GNNs, their lack of explainability hinders human trust and thus limits their application in safety-critical domains.
To mitigate this issue, recent efforts have explored identifying informative subgraphs that serve as either post-hoc or intrinsic explanations for the decisions made by GNNs.
Specifically, a line of post-hoc studies~\cite{ying2019gnnexplainer, luo2020parameterized, vu2020pgm} work on a pre-trained model and propose different combinatorial search methods for identifying the most influential subgraphs based on model predictions.
However, since these methods train another explanatory model to provide explanations, they may be disloyal to the original model, resulting in distorted attribution analysis.
In contrast to the post-hoc methods, the intrinsically interpretable ones endeavour to identify subgraphs during training and make reliable predictions guided by these subgraphs~\cite{ming2019interpretable, yu2020graph, yin2023train}.
The pioneering works, \textit{e.g.} GIB~\cite{ming2019interpretable} and GSAT~\cite{yu2020graph}, adopt the information bottleneck principle~\cite{wu2020graph} to constraint the information flow from input graph to prediction, ensuring the label-relevant graph components will be kept while the label-irrelevant ones are reduced.
Additionally, ProtGNN~\cite{zhang2022protgnn} learns representative subgraphs~(\textit{i.e.}, prototypes) from inputs by prototype learning~\cite{kolodner1992introduction} and makes predictions based on the similarity between new instances and prototypes. 
Unfortunately, the explanation graph is generated by an extra projection process based on the prototype embedding, which can introduce explanatory biases.

Graph-level tasks often necessitate global-level explanations to depict long-range dependencies and global interactions considering the whole graph~\cite{cao2015grarep, ding2023eliciting, lee2021learnable, yao2022trajgat}.
For example, in the case of protein molecules, enzymes are distinguished from other non-enzyme proteins by having fewer helices, more and longer loops, and tighter packing between secondary structures~\cite{stawiski2000predicting}. Identifying such global structural patterns often requires the collective participation of dozens or even hundreds of amino acids.
It is time-consuming to entail expert examination over the subgraph explanations of each node provided by previous subgraph-specific methods.
Beyond the node-wise representations of early GNNs, recent \emph{state-of-the-art} GNNs~\cite{wu2021representing, chen2022structure, kong2023goat, rampavsek2022recipe} have shifted the focus towards considering global interactions for graph-level tasks, enhancing the expressive power of GNNs by a large margin.
Hence, there exists a significant gap between local subgraph-specific explanations and global-level explanations, which are required by both graph-level tasks and advanced GNNs.

In this paper, we propose the Global Interactive Pattern (GIP) learning, a new interpretable graph classification task that approaches the problem from a global perspective.
This task poses two key challenges for existing techniques, namely, high computational complexity and diverse global patterns.
Firstly, the presence of a large number of nodes, along with their intricate connectivity, presents a significant challenge in modeling long-range dependencies and accurately extracting global interactions.
Simply extending subgraph-specific methods to identify global interactive patterns would result in exponentially increasing computational complexity. 
This is particularly true in real-world graphs where these patterns typically involve dozens or even hundreds of nodes.
Secondly, there exist multiple interactive patterns for graphs belonging to the same class.
Existing techniques either provide instance-level explanations or entail high costs for extracting graph patterns. 
Hence, it becomes crucial to identify representative and diverse patterns within an acceptable computational overhead for more comprehensive and accurate explanations.

To tackle these challenges, we explore an innovative framework for sloving GIP, by first performing compression of the graph and then identifying inter-cluster interactions in the coarsened graph instances, which we call interactive patterns, to determine the intrinsic explanations.
Specifically, the framework consists of two key modules: clustering assignment module and interactive pattern matching module.
First, in the clustering assignment module, we iteratively aggregate components with similar features or tight connections to form a cluster-level representation, and then extract global structure information based on the interactions between local structures, thus realizing the modeling of the global interactions while aggregating the information of local substructures.
Then, in the interactive pattern matching module, different from prior researches~\cite{zhang2022protgnn,dai2022towards} in graph pattern recognition that target at learning representative embeddings in hidden space, we define learnable interactive patterns in the form of graph structure to directly reveal the vital patterns in the graph level. Additionally, we introduce graph kernels as a measure of similarity between the coarsened graph and the interactive patterns, thereby propelling the learning and matching of interactive patterns based on the similarity.
Finally, with the similarity scores, a fully connected layer with softmax is applied to compute the output probabilities for each class.


In summary, the main contributions of our work are as follows:

\begin{itemize}[leftmargin=*]
\item We explore a novel interpretable graph classification task termed as Global Interactive Pattern~(GIP) learning, taking a step further from local subgraph explanation to global interactive patterns.
\item We propose a holistic framework for solving GIP, which achieves a double-win of high computational efficiency and accurate pattern discovery. By integrating learnable cluster constraints and graph prototypes, we can adaptively provide the decisions with reliable graph-level explanations.
\item 
Extensive experiments on both real-world and synthetic datasets demonstrate the effectiveness of our framework in achieving accurate prediction and valid explanation.
In addition, visualization of the explanations further demonstrates the superior capability of our framework in identifying global interactive patterns.
\end{itemize}

\begin{figure*}[ht]
  \includegraphics[width=\textwidth]{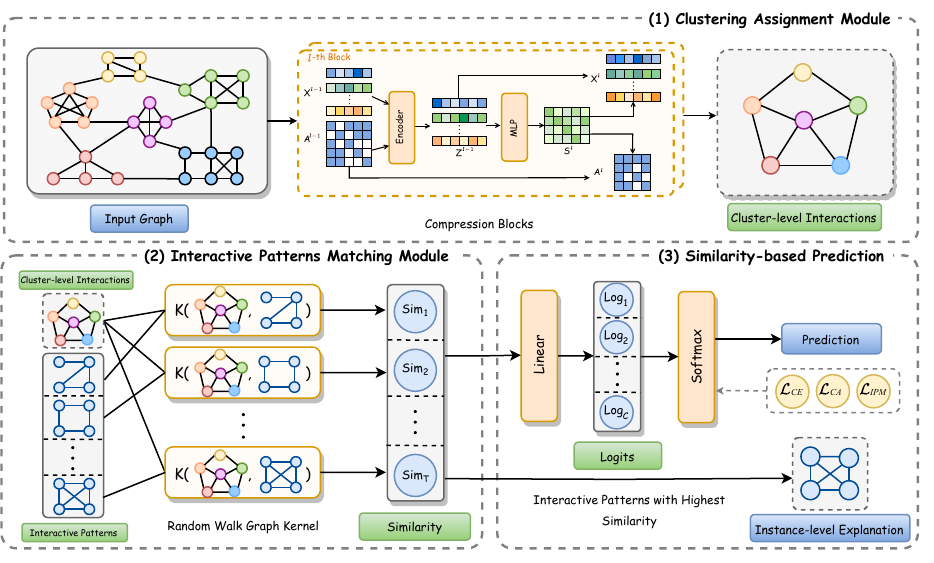}
  \caption{The architecture of our proposed two-stage framework for GIP.}
  \label{fig:fig1}
  \vspace{-0.3cm}
\end{figure*}

\section{Related Work}
\subsection{Graph Neural Networks}
Driven by the momentous success of deep learning, recently, a mass of efforts have been devoted to developing deep neural networks for graph-structured data~\cite{cao2016deep}. As one of the pioneer works, graph neural networks~(GNNs)~\cite{kipf2016semi,hamilton2017inductive,ying2018hierarchical,jing2021amalgamating} have demonstrated effectiveness in various real-world scenarios~\cite{jing2023deep,jing2022learning,zheng2022transition,wang2024spatiotemporal} such as traffic analysis~\cite{chen2019gated, zhang2021traffic}, drug generation~\cite{bongini2021molecular} and recommendation systems~\cite{, ying2018graph}.
Generally, classic GNN variants adopt the message-passing mechanism~\cite{gilmer2017neural} to update the embeddings of each node based on the calculated message set between the node and each of its neighbors.
Then, these node-wise representations are manipulated through concatenation or pooling operations to form graph-level representations for graph-level tasks.
Although this unique message-passing mechanism enables GNNs to fully leverage the relationships between nodes in graph structure, such GNNs may suffer from \emph{over-smoothing} due to repeated local aggregation and \emph{over-squashing} due to the exponential growth of computational cost with increasing model depth.
Recent years have witnessed many successful architectures that shift the focus towards considering global interactions for graph-level tasks.
These approaches~\cite{wu2021representing, chen2022structure, kong2023goat, rampavsek2022recipe} model long-range dependencies and global structures to facilitate a more comprehensive acquisition of the global information in graphs, thus enhancing the expressive power of the model.
Owing to the powerful representation capability, these GNNs have achieved state-of-the-art performance.

\subsection{Explainability of Graph Neural Networks}
Despite the great success of GNNs, their black-box nature undermines human trust, thereby hindering their application in high-stake domains.
To bolster understanding of GNNs and provide more credible evidence for decision-making, plenty of researches focus on the explainability of GNNs is emerging. Such studies concentrate on identifying vital subgraphs, offering intrinsic or post-hoc explanations for GNNs. 
The post-hoc explainable methods focus on designing different combinatorial search method to explore important subgraphs based on model outputs~\cite{pope2019explainability, ying2019gnnexplainer, luo2020parameterized, yuan2021explainability}.
As an initial endeavour, GNNExplainer~\cite{ying2019gnnexplainer} learns soft masks from edge and node features to identify pivotal subgraphs for explaining the prediction result.
Furthermore, PGExplainer~\cite{luo2020parameterized} employs a reparameterization trick to obtain approximated discrete masks instead of soft masks.
In addition, XGNN~\cite{yuan2020xgnn} generates representative subgraphs for different classes as model-level explanations.

Since these methods focus on providing post-hoc explanations for a trained GNN, they might fail to fit the original model precisely and generate biased explanation.
Though it would be preferable to design interpretable GNNs~\cite{chen2019looks, ming2019interpretable}, there are still limited efforts in this regard~\cite{dai2021towards, zhang2022protgnn, dai2022towards}. The goal of these methods is to identify subgraphs during training and make reliable predictions guided by subgraphs~\cite{ming2019interpretable, yu2020graph, yin2023train}.
GIB~\cite{ming2019interpretable} and GSAT~\cite{yu2020graph} adopt the information bottleneck principle~\cite{wu2020graph} to constraint the information flow from the input graph to the prediction, ensuring the label-relevant components will be kept while the label-irrelevant ones are reduced.
In addition, some existing works attempt to apply prototype learning for exploring important subgraphs from instances and make predictions based on the similarity between new instances and prototypes~\cite{zhang2022protgnn,dai2022towards}.
For example, ProtGNN~\cite{zhang2022protgnn} applies the Monte Carlo tree search~\cite{silver2017mastering} to identify subgraphs in the original graphs as prototypes, while PxGNN~\cite{dai2022towards} obtains prototypes from learnable prototype embeddings by a pre-trained prototype generator.

However, the aforementioned methods only provide one-side attribution analysis from a localized viewpoint, which may lead to under-representative explanations when higher-order node interactions or global graph structure play a pivotal role.
To address this issue, in this paper, we propose an interpretable scheme for graph classification called GIP, that explicitly extracts global interactive patterns to deliver graph-level explanations.

\section{Method}
In this section, we elaborate the details of the proposed framework for GIP.
First, in the clustering assignment module, we extract inter-cluster interactions from coarsened graph as global structural information.
Then, in the interactive pattern matching module, we match the coarsened graph with a batch of learnable interactive patterns based on the similarity calculated by the graph kernel.
Finally, with the similarity scores, the fully connected layer with softmax computes the probability distributions for each class.
The architecture of the proposed framework is shown in Figure~\ref{fig:fig1}.

\subsection{Preliminaries}
\subsubsection{Notations}
We denote an attributed graph with $N$ nodes by ${G}=(\mathbf{V}, \mathbf{X}, \mathbf{A})$, where $\mathbf{V}=\{v_1,...,v_N\}$ is the set of nodes in graph, $\mathbf{X} \in \mathbb{R}^{N \times d}$ is the matrix consisting of the $d$-dimensional feature vector of each node, $\mathbf{A} \in \{0, 1\}^{N \times N}$ is the adjacency matrix.
$\mathbf{A}_{ij} = 1$ if nodes $v_i$ and $v_j$ are connected; otherwise $\mathbf{A}_{ij} = 0$.

In this paper, we take graph classification as the target task.
Given a set of $M$ graphs $\mathcal{G} = \{{G}_1, {G}_2,..., {G}_M\}$, and each graph ${G}_m$ is associated with a ground-truth class label $y_m \in \mathcal{C}$, where $\mathcal{C}=\{1,2,...,C\}$ is the set of candidate labels. The graph classification task aims to learn a graph classifier that predicts the estimated label $\hat{y}_m$ for an input graph ${G}_{m}$.

\subsubsection{Graph Normalized Cut}
Graph normalized cut is an effective approach for realizing graph clustering. The goal is to construct a partition of the graph into $K$ sets, such that the sets are sparsely connected to each other while the internal structure of the sets exhibits high cohesion~\cite{rangapuram2014tight}. 
We formalize the objective of the $K$-way normalized cut as follows:
\begin{align}
  \mathop{\text{min}}_{\mathbf{V}_{1},...,\mathbf{V}_{K}} \frac{1}{K}\sum_{k=1}^{K}\frac{\text{cut}(\mathbf{V}_k, \overline{\mathbf{V}}_k)}{\text{vol}(\mathbf{V}_{k})},
\label{eq1}
\end{align}
where $\mathbf{V}_{k}$ represents the nodes belonging to cluster $k$, $\text{vol}(\mathbf{V}_{k})$ = $\sum_{i,j \in \mathbf{V}_{k}}\mathbf{A}_{ij}$ counts the number of edges within cluster $k$, and $\text{cut}(\mathbf{V}_k, \overline{\mathbf{V}}_k)$ = $\sum_{i \in \mathbf{V}_{k}, j \in \mathbf{V} \backslash {\mathbf{V}_{k}}}\mathbf{A}_{ij}$ counts the edges between the nodes in cluster $k$ and the rest of the graph~\cite{shi2000normalized}.
Let $\mathbf{P} \in \{0,1\}^{N \times K}$ be the cluster assignment matrix, where $K$ denotes the number of target clusters and $\mathbf{P}_{ij}=1$ when node $i$ belongs to cluster $j$.
The objective function of the normalized cut can be further defined according to the derivation in~\cite{duong2023deep,chen2022finc}:
\begin{align}
  & \mathop{\text{min}}_{\mathbf{P} \in \{0,1\}^{N \times K}}\frac{1}{K}\sum_{k=1}^{K}\frac{\mathbf{P}_k^{T}\mathbf{L}\mathbf{P}_k}{\mathbf{P}_k^{T}\mathbf{D}\mathbf{P}_k} \\
 = & \mathop{\text{min}}_{\mathbf{P} \in \{0,1\}^{N \times K}}{\frac{1}{K} \cdot \text{Tr}}(\frac{\mathbf{P}^{T}\mathbf{L}\mathbf{P}}{\mathbf{P}^{T}\mathbf{D}\mathbf{P}}),
 \label{eq2}
\end{align}
where $\mathbf{P}_k$ represents the $k$-th column in $\mathbf{P}$, $\mathbf{D}$ is the corresponding degree matrix, and $\mathbf{L}=\mathbf{D}-\mathbf{A}$ is the graph Laplacian matrix.

The optimization problem is NP-hard because the clustering assignment matrix $\mathbf{P}$ takes discrete values~\cite{shi2003multiclass}.
Therefore, following the traditional approach of solving the probabilistic approximation of the $K$-way normalized cut~\cite{duong2023deep,chen2022finc}, we perform a continuous relaxation for $\mathbf{P}$ such that it satisfies $\mathbf{P}_{ij} \in [0,1]$ and $ \forall i, \sum_j \mathbf{P}_{ij} = 1$.

\subsubsection{Random Walk Graph Kernel}
Random walk graph kernel is a kind of kernel function for graph similarity evaluation, whose core idea is to compute the similarity of two input graphs by counting the number of common paths in the two graphs. $R$-step random walk means that the length of paths formed by the random walk does not exceed $R$.
To efficiently compute the random walk kernel, we follow the generalized framework of computing walk-based kernel~\cite{vishwanathan2006fast}, and use the direct product graph for equivalence calculation.

Given two graphs ${G} = (\mathbf{V}, \mathbf{X}, \mathbf{A})$ with $N$ nodes and ${G}^{\prime} = (\mathbf{V}^{\prime}, \mathbf{X}^{\prime}, \mathbf{A}^{\prime})$ with $N^{\prime}$ nodes, the direct product graph ${G}_{\times} =(\mathbf{V}_{\times}, \mathbf{X}_{\times}, \mathbf{A}_{\times})$ is a graph with $NN^{\prime}$ nodes, each representing a pair of nodes from $G$ and $G^{\prime}$. 
The adjacency matrix $\mathbf{A}_{\times}$ is equal to the Kronecker product of the adjacency matrices of $G$ and $G^{\prime}$, that is $\mathbf{A}_{\times}=\mathbf{A}\otimes\mathbf{A}^{\prime}$~\cite{borgwardt2006fast}.
The attribute of node $(v,v^{\prime})$ in $G_{\times}$ is calculated based on the attribute of node $v$ in $G$ and node $v^{\prime}$ in $G^{\prime}$, i.e. $\mathbf{X}_{{\times}_{(v,v^{\prime})}}=\mathbf{X}_{v}\mathbf{X}^{\prime T}_{v^{\prime}}$.
Performing a random walk on the direct product graph ${G}_{\times}$ is equivalent to performing the simultaneous random walks on graphs ${G}$ and ${G}^{\prime}$. Therefore, The $R$-step random walk kernel for attributed graphs~\cite{feng2022kergnns} can be calculated as:
\begin{align}
  K({G}, {G}^{\prime}) = \sum_{r=0}^{R}K_{r}({G}, {G}^{\prime})
\label{eq4}
\end{align}

\begin{align}
  K_{r}({G}, {G}^{\prime}) &= \sum_{i,j=1}^{|\mathbf{V}_{\times}|}\mathbf{X}_{\times_i}\mathbf{X}_{\times_j}{[\mathbf{A}_{\times}^{r}]}_{ij}
\label{eq5}
\end{align}
where $\mathbf{X}_{\times_i}$ denotes the feature of $i$-th nodes in ${G}_\times$ and the $(i, j)$-th element of $\mathbf{A}_{\times}^{r}$ represents the number of common walks of length $r$ between the $i$-th and $j$-th node in ${G}_{\times}$.

\subsection{Clustering Assignment Module}
\label{Section3.2}
In this module, the underlying idea of our approach stems from related work on graph pooling~\cite{ying2018hierarchical}, which progressively creates coarser versions to represent cluster-level interactions by applying a series of compression blocks to the input graph.
In each compression block, we first obtain the embedding vector $\mathbf{Z} \in \mathbb{R}^{N \times d^{\prime}}$ of nodes by encoder, which can be any model,
and we apply GCN~\cite{kipf2016semi} as encoder for implementation.

\begin{align}
  \widetilde{\mathbf{A}} &= \hat{\mathbf{D}}^{-\frac{1}{2}}\hat{\mathbf{A}}\hat{\mathbf{D}}^{-\frac{1}{2}}, \\
  \mathbf{Z} &= f(\{\mathbf{X}, \widetilde{\mathbf{A}}\}; \Theta_\text{GCN}),
\end{align}
where $\hat{\mathbf{A}} = \mathbf{A} + \mathbf{I}_{N}$ is the adjacency matrix with added self-loop, $\hat{\mathbf{D}}$ is the degree matrix of $\hat{\mathbf{A}}$, and $\Theta_\text{GCN}$ are parameters of the encoder.

Then, we divide the original input graph into the cluster-level representation based on the generated node embeddings in a trainable manner.
Specifically, we define a trainable cluster assignment matrix $\mathbf{S}$ to map each node to a corresponding cluster, and each entry $\mathbf{S}_{ij}$ represents the probability of node $i$ belonging to cluster $j$.
Considering that the similarity of node features can affect clustering assignment to some extent, node feature embedding is incorporated into the learning process of $\mathbf{S}$. We take $\mathbf{Z}$ as input and use a multi-layer perceptron~(MLP) with softmax on the output layer to compute $\mathbf{S}$:
\begin{align}
  \mathbf{S} = \text{Softmax}\Big (\text{MLP}\big (\mathbf{Z}; \Theta_{\text{MLP}_1} \big)\Big ),
\end{align}
where $\mathbf{S}$ satisfies $\mathbf{S}_{ij} \in [0,1]$ and $\forall i \sum_j \mathbf{S}_{ij} = 1$, $\Theta_{\text{MLP}_1}$ denotes the learnable parameters in the MLP.

Unlike the unconstrained learning process in~\cite{ying2018hierarchical}, we aim to impose constraints on $\mathbf{S}$ in order to obtain clustering assignment results that better reflect the clustering characteristics of nodes in the real-world graphs.
First, we optimize the learning of $\mathbf{S}$ by minimizing an unsupervised loss term $\mathcal{L}_{clu}$, which defined on a relaxation formula that approximates the $K$-way normalized cut~(\ref{eq2}):
\begin{align}
  \mathcal{L}_{clu} = {\frac{1}{K} \cdot \text{Tr}}(\frac{\mathbf{S}^{T}\mathbf{L}\mathbf{S}}{\mathbf{S}^{T}\mathbf{D}\mathbf{S}}),
\end{align}
where $\mathbf{D}$ is the corresponding degree matrix, and $\mathbf{L}=\mathbf{D}-\mathbf{A}$ is the graph Laplacian matrix.
However, without additional constraints on the assignment matrix $\mathbf{S}$, cluster assignment may fall into a local optimal solution: assigning all nodes to the same cluster.
Hence, we introduce an balanced loss term $\mathcal{L}_{bal}$ to encourage more balanced and discrete clusters:
\begin{align}
  \mathcal{L}_{bal} = \frac{\sqrt{K}}{N}||\sum_{i=1}^{N}\mathbf{S}_{i}||_{F}-1,
\end{align}
where $||\cdot||_{F}$ indicates the Frobenius norm, $N$ is the number of nodes and $K$ is the number of target clusters.

In summary, the optimization objective of this module can be expressed as:
\begin{align}
\label{eq:loss1}
  \mathcal{L}_{CA} = \alpha_{1} \mathcal{L}_{clu} + \alpha_{2} \mathcal{L}_{bal},
\end{align}
where $\alpha_{1}$ and $\alpha_{2}$ control the ratio of the loss terms.

Assuming the input adjacency matrix in the $\ell$-th compression block is $\mathbf{A}^{\ell-1}$, the input node embedding matrix is $\mathbf{Z}^{\ell-1}$, and the computed clustering assignment matrix is $\mathbf{S}^{\ell}$, we can generate a new coarsened adjacency matrix $\mathbf{A}^{\ell}$ and a new embedding matrix $\mathbf{X}^{\ell}$ for next compression block. Specifically, we apply the following two equations:
\begin{align}
  \mathbf{X}^{\ell} &= {\mathbf{S}^{\ell}}^{T}\mathbf{Z}^{\ell-1}\in\mathbb{R}^{N^{\ell}\times d},\\
  \mathbf{A}^{\ell} &= {\mathbf{S}^{\ell}}^{T}\mathbf{A}^{\ell-1}{\mathbf{S}^{\ell}}\in\mathbb{R}^{N^{\ell}\times N^{\ell}},
\end{align}
where $N^{\ell}$ denotes the number of target clusters in $\ell$-th block and $d$ denotes dimension of node features.
By stacking compression blocks, we can obtain $\mathbf{A}^{L}$ and $\mathbf{X}^{L}$ for cluster-level representation ${CG}$, where $L$ is the number of compression blocks.
Considering the impact of the enormous edges in the coarsened graph, we propose to filter the edges.
Specifically, we define the matrix $\mathbf{Mask} \in \{0,1\}^{N^{L} \times N^{L}}$ to filter the edges in the coarsened graph, where $N^{L}$ is the number of nodes in the coarsened graph.
If $\mathbf{A}^{L}_{ij}$ exceeds threshold ${\delta}_1$, the element at the corresponding position in $\mathbf{Mask}$ is set to 1, otherwise it is set to 0:
\begin{align}
  \mathbf{Mask}_{ij}=
  \left\{
	\begin{aligned}
  &1, \text{if}~\mathbf{A}^{L}_{ij} > {\delta}_1;\\
  &0, \text{else},
	\end{aligned}
  \right.
\end{align}
Thus, we obtain the filtered adjacency matrix ${\mathbf{A}^{L}}^{\prime}=\mathbf{A}^{L} \odot \mathbf{Mask}$ for cluster-level representation, where $\odot$ is the element-wise product.

\subsection{Interactive Patterns Matching Module}
In this module, we aim to learn representative inter-cluster structures and interactions for each class, which we call interactive patterns, to give accurate predictions and reliable explanations.

First, we define a total of $T$ learnable interactive patterns, i.e. $\mathcal{P}=\{P_1, P_2, ..., P_T\}$, and allocate them evenly to $C$ classes.
In order to provide a more understandable explanation, we define each interactive pattern $P_t$ as a combination of the following two parts:
(i) randomly initialized feature matrix $\mathbf{X}^{P_t}$ with pre-defined size;
(ii) the topology $\mathbf{A}^{P_t}$ generated from the feature matrix, and the generation process of $\mathbf{A}^{P_t }$ is defined as follows:
\begin{align}
    \mathbf{A}^{P_i}_{ij} = \sigma \Big (\text{MLP}\big ([\mathbf{X}^{P_t}_i;\mathbf{X}^{P_t}_j];\Theta_{\text{MLP}_2}\big)\Big)
\end{align}
where $\sigma(\cdot)$ is the Sigmoid function, $\Theta_{\text{MLP}_2}$ is trainable parameters of MLP, $[\cdot;\cdot]$ is concatenation operation, $\mathbf{X}^{P_t}_i$ and $\mathbf{X}^{P_t}_j$ are features of nodes in interactive pattern.   
Therefore, the generated interactive patterns can be directly used for explanation without the need for additional graph projection or graph generation processes~\cite{zhang2022protgnn,dai2022towards}.

Then, for the coarsened graph ${CG}$ and interactive pattern $P_t$, we propose to calculate their similarity through graph kernels~\cite{borgwardt2005shortest, kashima2003marginalized}. The choice of graph kernels can be changed according to the actual application scenario. Here, we choose the $R$-step random walk graph kernel~\cite{gartner2003graph, vishwanathan2006fast} which compares random walks up to length $R$ in two graphs. Then, the similarity between the coarsened graph ${CG}$ and the interactive pattern ${P_t}$ can be expressed as: 
\begin{align}
  \text{sim}({CG}, {P_t}) = K({CG}, {P_t}),
\end{align}
where $K({CG}, {P_t})$ is calculated by equations~(\ref{eq4}) and~(\ref{eq5}).

Considering the desired representativeness of the interactive patterns for their corresponding classes, we suppose that the learning objective of interactive patterns is to encourage each coarsened graph to approach the interactive patterns belonging to the same class, while moving away from the interactive patterns belonging to other classes.
To achieve this, we introduce the multi-similarity loss~\cite{wang2019multi} to constrain learning of patterns:
\begin{align}
\begin{aligned}
  \mathcal{L}_{mul}=\frac{1}{M}\sum_{m=1}^{M} & \Big(\frac{1}{{\gamma}_{1}}\log\big(1+\sum_{P_i \in \mathbf{Pos}_m}e^{\gamma_{1}(d_{mi}-\lambda)}\big) \\
  &+\frac{1}{\gamma_{2}}\log\big(1+\sum_{P_i \in \mathbf{Neg}_m}e^{-\gamma_{2}(d_{mi}-\lambda)}\big)\Big)
\end{aligned}
\end{align}
where $\mathbf{Pos}_m$ denotes the set of interactive patterns belonging to the same class as the coarsened graph $CG_{m}$, $\mathbf{Neg}_m$ denotes the set of interactive patterns apart from these, $d_{mi}$ denotes the distance between coarsened graph ${CG}_m$ and interactive pattern $P_i$, $\gamma_{1}$ and $\gamma_{2}$ control the contributions of different items, and $\lambda$ represents the margin which controls the distribution range of interactive patterns belonging to the certain class.
For the computation of $d_{mi}$, we apply the distance in kernel space~\cite{scholkopf2000kernel}:
\begin{align}
  d_{mi} = \sqrt{\frac{1}{2}\Big(K\big({CG}_m, {CG}_m\big)+K\big(P_i,P_i\big)\Big)-K\big({CG}_m, P_i\big)}
\end{align}

Additionally, we encourage diversity in interactive patterns by adding the diversity loss, which penalizes interactive patterns that are too close to each other:
\begin{align}
  \mathcal{L}_{div} = \sum_{c=1}^{C}\sum_{{P_i}, {P_j}\in \mathcal{P}_{c}}\text{max}\Big(0, sim\big({P_i}, {P_j}\big)-{\delta}_{2}\Big)
\end{align}
where $\mathcal{P}_{c}$ denotes the interactive patterns belonging to class $c$ and ${\delta}_2$ is the threshold for similarity measurement.
\begin{align}
\label{eq:loss2}
  \mathcal{L}_{IPM} = \alpha_{3}\mathcal{L}_{mul}+\alpha_{4}\mathcal{L}_{div}
\end{align}
where $\alpha_{3}$ and $\alpha_{4}$ control the ratio of the loss terms.

\subsection{Interpretable Classification with interactive patterns}
\subsubsection{Classification and Learning Objective}
Finally, the $T$ similarity scores between the coarsened graph and each interactive pattern are fed into the fully connected layer to obtain the output logits. Then, the logits processed with softmax to yield the probability distribution $h_i$ for a given graph $G_i$.
To ensure the accuracy of the proposed framework, we apply a cross-entropy loss to leverage the supervision from the labeled set:
\begin{align}
  \mathcal{L}_{CE} = \frac{1}{M}\sum_{i=1}^{M}\text{CrsEnt}(h_i,y_i)
\end{align}
where $y_i$ is the true label of input graph.
To sum up, the objective function we aim to minimize is:
\begin{align}
\label{eq:loss3}
  \mathcal{L} = \mathcal{L}_{CE} + \beta_1\mathcal{L}_{CA}+\beta_2\mathcal{L}_{IPM}
\end{align}
where $\mathcal{L}_{CA}$ and $\mathcal{L}_{IPM}$ are loss terms of the clustering assignment module and interactive patterns matching module, $\beta_1$ and $\beta_2$ control the contribution of these loss terms.

\subsubsection{Explainability}
From the class perspective, the learned interactive patterns $\mathcal{P}$ reveal the cluster-level interaction characteristics of the graphs in each class. 
From the instance perspective, for the test graph ${G}_{t}$, we can identify the most similar interactive pattern in class $\hat{y}_{t}$ with ${G}_{t}$ as the instance-level explanation:
\begin{align}
\hat{{G}_{t}^{*}} = \mathop{\arg\max}\limits_{{P_i}\in \mathcal{P}^{\hat{y}_{t}}} \, \text{sim}({G}_{t},{P_i})
\end{align}
where $\mathcal{P}^{\hat{y}_{t}}$ is the set of interactive patterns belonging to class $\hat{y}_{t}$.
Since the prediction of ${G}_{t}$ is based on several patterns, the instance-level explanation can be several similar patterns in class $\hat{y}_{t}$, thereby bringing deeper insights into the graph itself.

\begin{table*}[!t]
  \caption{Comparison of different methods in terms of classification accuracy~(\%) and F1 score~(\%). Baselines include widely used GNNs, interpretable GNNs and post-hoc explainable GNNs. The datasets include real-world datasets and synthetic datasets. \textbf{Bold} and \underline{underline} denote the best and the second-best results, respectively.}
  \label{tab:tabel2}
  \resizebox{\linewidth}{!}{
\begin{tabular}{@{}cccccccccccccc@{}}
\toprule
\textbf{Datasets}                    & \textbf{Metrics} & \textbf{GCN}     & \textbf{DGCNN}                                & \textbf{Diffpool}                             & \textbf{RWNN}                                                     & \textbf{GraphSAGE}                            & \textbf{ProtGNN} & \textbf{KerGNN}                            & \textbf{$\pi$-GNN} & \textbf{GIB}                                  & \textbf{GSAT}                                 & \textbf{CAL}                                  & \textbf{Ours}                                 \\ \midrule
\multirow{2}{*}{\textbf{ENZYMES}}    & \textbf{Acc.}     & $57.81 \pm 0.73$ & $58.68 \pm 2.74$                              & $59.99 \pm 2.35$                              & $56.94 \pm 0.87$                                                  & \multicolumn{1}{c|}{$56.33 \pm 0.88$}                              & $55.00 \pm 2.36$ & $57.68 \pm 3.79$                           & $57.85 \pm 0.66$   & $46.17 \pm 5.11$                              & \underline{$60.55 \pm 1.87$} & \multicolumn{1}{c|}{$60.03 \pm 5.40$}                              & $\mathbf{60.76 \pm 2.61}$    \\
                                     & \textbf{F1}      & $49.00 \pm 0.54$ & $52.77 \pm 2.46$                              & $\mathbf{58.89 \pm 2.64}$    & $49.91 \pm 8.19$                                                  & \multicolumn{1}{c|}{$42.64 \pm 1.46$}                              & $45.17 \pm 1.46$ & $49.90 \pm 1.47$                           & $48.74 \pm 0.74$   & $30.84 \pm 1.16$                              & $53.34 \pm 0.56$                              & \multicolumn{1}{c|}{$57.17 \pm 7.41$}                              & \underline{$58.45 \pm 3.41$} \\
\rowcolor{gray!15}
       & \textbf{Acc.}     & $75.92 \pm 3.63$ & $77.53 \pm 1.33$                              & $\mathbf{80.03 \pm 1.02}$    & $77.42 \pm 2.16$                                                  & \multicolumn{1}{c|}{\underline{$79.61 \pm 4.60$}} & $78.34 \pm 0.61$ & $74.66 \pm 1.39$                           & $78.63 \pm 1.12$   & $76.49 \pm 2.54$                              & $73.60 \pm 1.07$                              & \multicolumn{1}{c|}{$77.25 \pm 3.62$}                              & $79.52 \pm 0.50$                              \\
\rowcolor{gray!15}
 \multirow{-2}{*}{\textbf{D\&D}}                                & \textbf{F1}      & $72.08 \pm 0.33$ & $68.54 \pm 0.86$                              & $73.88 \pm 5.27$                              & \multicolumn{1}{l}{\underline{$76.14 \pm 1.87$}} & \multicolumn{1}{c|}{$\mathbf{78.62 \pm 8.86}$}    & $73.29 \pm 2.87$ & $66.29 \pm 2.47$                           & $71.14 \pm 2.13$   & $68.26 \pm 0.98$                              & $66.16 \pm 4.82$                              & \multicolumn{1}{c|}{$66.65 \pm 5.13$}                              & $73.67 \pm 1.71$                              \\
\multirow{2}{*}{\textbf{PROTEINS}}   & \textbf{Acc.}     & $79.05 \pm 1.17$ & $76.85 \pm 2.74$                              & \underline{$79.73 \pm 0.64$} & $74.43 \pm 1.20$                                                  & \multicolumn{1}{c|}{$79.04 \pm 1.03$}                              & $78.22 \pm 0.61$ & $78.15 \pm 2.21$                           & $73.34 \pm 0.64$   & $74.90 \pm 5.10$                              & $77.67 \pm 1.59$                              & \multicolumn{1}{c|}{$75.20 \pm 3.59$}                              & $\mathbf{79.84 \pm 0.81}$    \\
                                     & \textbf{F1}      & $70.97 \pm 2.89$ & $73.19 \pm 4.74$                              & $\mathbf{77.95 \pm 0.77}$    & $72.34 \pm 1.28$                                                  & \multicolumn{1}{c|}{$67.76 \pm 2.84$}                              & $73.79 \pm 2.87$ & $72.13 \pm 1.40$                           & $66.17 \pm 4.64$   & $71.11 \pm 0.18$                              & $72.86 \pm 0.96$                              & \multicolumn{1}{c|}{$66.02 \pm 3.67$}                              & \underline{$74.50 \pm 2.84$} \\
\rowcolor{gray!15}
& \textbf{Acc.}     & $74.32 \pm 8.10$ & $56.34 \pm 0.77$                              & $83.90 \pm 9.70$                              & $86.47 \pm 0.39$                                                  & \multicolumn{1}{c|}{$72.10 \pm 4.30$}                              & $81.67 \pm 2.36$ & $72.66 \pm 0.94$                           & $91.18 \pm 0.28$   & $91.04 \pm 6.40$                              & $\mathbf{94.42 \pm 0.92}$    & \multicolumn{1}{c|}{$88.92 \pm 8.37$}                              & \underline{$92.13 \pm 3.26$} \\
\rowcolor{gray!15}
\multirow{-2}{*}{\textbf{MUTAG}}  & \textbf{F1}      & $65.33 \pm 4.60$ & $47.35 \pm 0.67$                              & $69.99 \pm 1.10$                              & \underline{$84.69 \pm 0.18$}                     & \multicolumn{1}{c|}{$69.81 \pm 2.20$}                              & $62.69 \pm 3.81$ & $61.39 \pm 1.89$                           & $77.51 \pm 1.95$   & $80.64 \pm 1.13$                              & $81.75 \pm 0.21$                              & \multicolumn{1}{c|}{$84.36 \pm 7.22$}                              & $\mathbf{87.27 \pm 2.27}$    \\
\multirow{2}{*}{\textbf{COLLAB}}     & \textbf{Acc.}     & $72.35 \pm 1.57$ & $73.27 \pm 1.39$                              & $73.53 \pm 1.48$                              & $72.37 \pm 1.32$                                                  & \multicolumn{1}{c|}{$72.63 \pm 1.48$ }                             & $69.26 \pm 0.86$ & $75.39 \pm 1.78$                           & $74.11 \pm 1.23$   & $73.17 \pm 1.60$                              & $76.89 \pm 2.83$                              & \multicolumn{1}{c|}{$\mathbf{79.08 \pm 1.94}$}    & \underline{$77.72 \pm 2.31$} \\
                                     & \textbf{F1}      & $63.66 \pm 4.91$ & \underline{$69.51 \pm 0.84$} & $69.29 \pm 0.44$                              & $67.42 \pm 2.04$                                                  & \multicolumn{1}{c|}{$62.84 \pm 2.31$}                              & $68.92 \pm 1.14$ & $\mathbf{70.21 \pm 2.09}$ & $65.98 \pm 3.27$   & $60.54 \pm 2.52$                              & $64.15 \pm 3.26$                              & \multicolumn{1}{c|}{$61.52 \pm 6.80$}                              & $67.34 \pm 1.73$                              \\
\rowcolor{gray!15}
& \textbf{Acc.}     & $81.21 \pm 1.08$ & $79.33 \pm 2.42$                              & $79.17 \pm 3.53$                              & $80.50 \pm 1.65$                                                  & \multicolumn{1}{c|}{$79.77 \pm 1.12$}                              & $78.11 \pm 1.05$ & $79.29 \pm 0.77$                           & $80.53 \pm 1.49$   & $79.87 \pm 0.78$                              & $81.17 \pm 0.86$                              & \multicolumn{1}{c|}{\underline{$81.94 \pm 1.07$}} & $\mathbf{82.51 \pm 1.18}$    \\
\rowcolor{gray!15} \multirow{-2}{*}{\textbf{GraphCycle}} & \textbf{F1}      & $72.95 \pm 1.03$ & $74.23 \pm 3.24$                              & $69.77 \pm 4.86$                              & $\mathbf{78.52 \pm 2.76}$                        & \multicolumn{1}{c|}{$71.16 \pm 2.56$}                              & $70.82 \pm 2.25$ & $71.82 \pm 0.61$                           & $75.98 \pm 4.87$   & $73.43 \pm 2.17$                              & $74.12 \pm 0.38$                              & \multicolumn{1}{c|}{$75.83 \pm 3.24$}                              & \underline{$77.91 \pm 5.73$} \\ 
\multirow{2}{*}{\textbf{GraphFive}}  & \textbf{Acc.}     & $58.96 \pm 2.28$ & $57.38 \pm 3.50$                              & $54.93 \pm 2.27$                              & $58.79 \pm 1.51$                                                  & \multicolumn{1}{c|}{$59.49 \pm 0.46$}                              & $56.57 \pm 3.38$ & $57.94 \pm 0.54$                           & $59.39 \pm 0.19$   & \underline{$59.71 \pm 0.65$} & $58.77 \pm 0.54$                              & \multicolumn{1}{c|}{$57.35 \pm 0.52$}                              & $\mathbf{60.40 \pm 1.75}$    \\
                                     & \textbf{F1}      & $53.83 \pm 0.91$ & $53.31 \pm 4.85$                              & $52.36 \pm 1.32$                              & {$53.65 \pm 1.01$}                                                  & \multicolumn{1}{c|}{$51.02 \pm 0.48$}                              & $54.25 \pm 3.51$ & $49.74 \pm 0.04$                           & $52.29 \pm 0.78$   & $\mathbf{58.72 \pm 0.63}$    & $54.54 \pm 2.39$                              & \multicolumn{1}{c|}{$51.63 \pm 1.41$}                              & \underline{$55.89 \pm 2.52$} \\ \bottomrule
\end{tabular}
  }
\end{table*}

\section{Experiments}
\subsection{Experimental Settings}
\subsubsection{Datasets}
In the experiment, we use five real-world datasets with different characteristics (e.g., size, density, etc.) for graph classification.
Additionally, to better demonstrate the explainability provided by our framework, we design two synthetic datasets.
The specific information of the datasets is as follows:
\begin{itemize}[leftmargin=*]
  \item {Real-world Datasets}: To probe the effectiveness of our framework in diffrent domains, we use protein datasets including ENZYMES, PROTEINS~\cite{feragen2013scalable}, D\&D~\cite{dobson2003distinguishing}, molecular dataset MUTAG~\cite{wu2018moleculenet} and scientific collaboration dataset COLLAB~\cite{yanardag2015deep}.
  The statistics of the datasets are presented in Appendix~\ref{appendix_1.1}.
  \item {Synthetic Datasets}: To better demonstrate the interpretability of our framework, we design two synthetic datasets: GraphCycle and GraphFive.
  Their labels are based on the interactive patterns between local structures. GraphCycle consists of two classes: Cycle and Non-Cycle, while GraphFive consists of five classes: Wheel, Grid, Tree, Ladder, and Star.
  The specific implementation details are presented in Appendix~\ref{appendix_1.1}.
\end{itemize}

\subsubsection{Baselines}
We extensively compare our framework with the following three types of baselines:
\begin{itemize}[leftmargin=*]
  \item {Widely Used GNNs}: We compare the prediction performance with the powerful GNN models including GCN~\cite{kipf2016semi}, DGCNN~\cite{wang2019dynamic}, Diffpool~\cite{ying2018hierarchical}, RWNN~\cite{nikolentzos2020random} and GraphSAGE~\cite{hamilton2017inductive}.
  \item {Post-hoc Explainable GNNs}: We compare the explanation performance with the post-hoc explainable methods including GNNExplainer~\cite{ying2019gnnexplainer}, SubgraphX~\cite{yuan2021explainability} and XGNN~\cite{yuan2020xgnn}.
  \item {Interpretable GNNs}: We compare the prediction and explanation performance with interpretable models including ProtGNN~\cite{zhang2022protgnn}, KerGNN~\cite{feng2022kergnns}, $\pi$-GNN~\cite{yin2023train}, GIB~\cite{yu2020graph}, GSAT~\cite{miao2022interpretable} and CAL~\cite{sui2022causal}.
\end{itemize}
More experimental settings will be presented in Appendix~\ref{appendix_1.2}
\subsection{Quantitative Analysis}
To validate the effectiveness of our framework, we first compare it with the baselines in terms of prediction and explanation performance on several graph classification datasets.

\subsubsection{Prediction Performance}
To demonstrate the effectiveness of our approach in providing accurate predictions, we choose classification accuracy and F1 scores as evaluation metrics, and compare them with widely used GNNs and interpretable GNNs on both real-world and synthetic datasets.
We apply three independent runs and report the average results along with the standard deviations in Table~\ref{tab:tabel2}. From the Table~\ref{tab:tabel2}, we can observe that:
\begin{itemize}[leftmargin=*]
  \item \textbf{Our framework achieves superior prediction performance compared to most of widely used GNNs.}
  Specifically, in terms of classification accuracy, our framework outperforms widely used GNNs on six of the seven datasets. 
  Particularly on MUTAG, our framework outperforms widely used models by 5.66\%\textasciitilde 35.79\%.
  Furthermore, for the dataset in which our framework lagged behind (D\&D), our framework only falls behind by $0.5\%$ compared to the best-performing widely used model.
  For the F1 score metric, our framework surpasses all widely used baselines in two of the seven datasets. Additionally, it achieves second-best performance in three datasets. In the remaining two datasets, it also performs comparably to most of widely used baselines.
  \item \textbf{Our framework significantly outperforms the leading interpretable GNNs in prediction performance.}
  On four of the seven datasets, our framework exceeds previous interpretable methods in terms of both accuracy and F1 score.
  On the remaining three datasets, although its accuracy/F1 score is slightly lower than the best-performing interpretable method, it still maintains the best performance in another metric.
  This demonstrates that our framework can consistently learn high-quality patterns for accurate predictions on different datasets; while simply selecting subgraphs might result in sub-optimal results.
\end{itemize}

\begin{figure*}[!ht]
  \includegraphics[width=\textwidth]{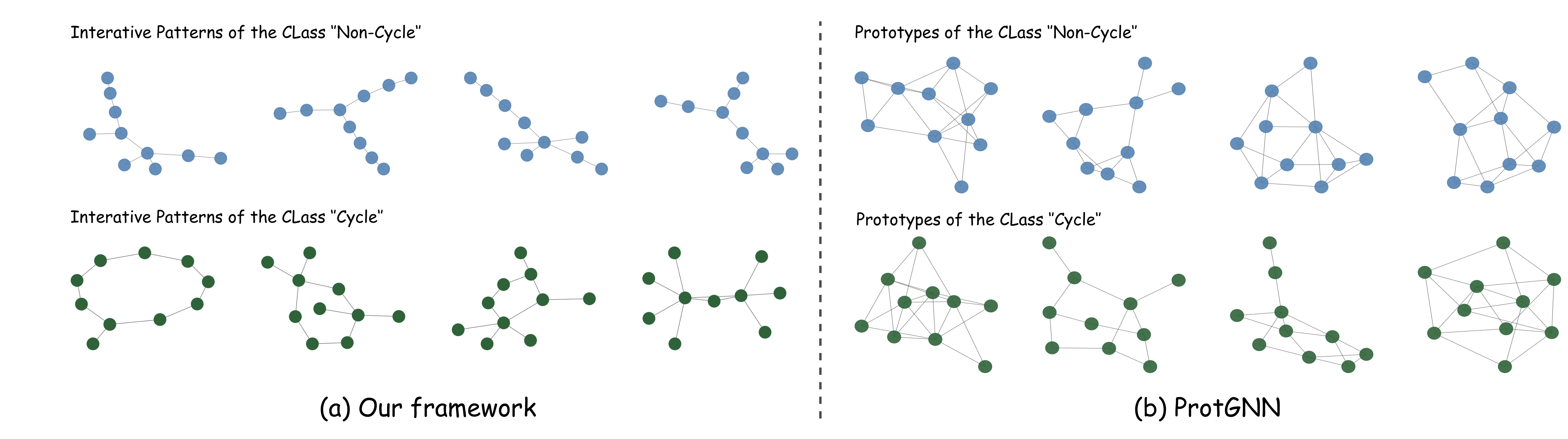}
  \caption{Explanations generated by our framework and ProtGNN on the GraphCycle dataset.}
  \label{fig:fig4}
\end{figure*}

\subsubsection{Explanation Performance}
We further compare the explanation performance of our method with that of interpretable methods and post-hoc explainable methods with three evaluation metrics, including explanation accuracy, consistency and silhouette score. We perform three independent runs and report the average results.

\begin{itemize}[leftmargin=*]
  \item \textbf{Explanation Accuracy.}
  We use trained GNNs to predict the explanations produced by different methods and take the confidence score of the prediction as the accuracy of the explanation~\cite{dai2022towards,li2022survey}.
  We compare our framework with interpretable methods and post-hoc explainable methods, the results are shown in Figure~\ref{tab:tabel3}.
  Compared to previous interpretable methods, our method exhibits the highest explanation accuracy in five out of seven datasets, and achieves the second-best performance in the remaining dataset.
  Compared to post-hoc explainable methods, our method also achieves the highest explanation accuracy on most datasets.
  
  \item \textbf{Consistency.}
  In the two synthetic datasets, we calculate the similarity between the explanations produced by different methods and the ground-truth.
  Here, we use the normalized results of random walk graph kernel as the measure of similarity.
  The results are presented in Table~\ref{tab:tabel4}.
  Our framework outperforms other baselines by a significant margin across all datasets. This indicates that our framework can provide more accurate explanations.
  
  \item \textbf{Silhouette Score.}
  High-quality interactive patterns can tightly cluster instances in dataset.
  Therefore, we use generated interactive patterns as centers to assign each graph to the nearest interactive pattern and then calculate the silhouette scores~\cite{rousseeuw1987silhouettes} to evaluate the compactness and separability of the clusters.
  We compare our method with another prototype-based approach ProtGNN, and the results are shown in Table~\ref{tab:tabel5}.
  Our method consistently achieves better performance on all datasets, which further demonstrates that our framework can obtain more representative patterns.
  
\end{itemize}

\begin{table}[!t]\Huge
    \renewcommand\arraystretch{1.4}
    \caption{Comparison of different methods in terms of explanation accuracy.}
    \label{tab:tabel3}
    \resizebox{\linewidth}{!}{
        \begin{tabular}{@{}cccccccc@{}}
        \toprule
        \textbf{Method}       & \textbf{ENZYMES} & \textbf{D\&D}      & \textbf{PROTEINS} & \textbf{MUTAG}   & \textbf{COLLAB}  & \textbf{GraphCycle} & \textbf{GraphFive} \\ \midrule
        \textbf{ProtGNN}      & $\mathbf{86.52 \pm 2.26}$ & $78.27 \pm 2.59$ & $67.34 \pm 3.89$  & $69.74 \pm 2.98$ & $78.53 \pm 3.41$ & $80.52 \pm 1.82$    & $71.48 \pm 1.69$   \\
        \textbf{KerGNN}       & $62.95 \pm 2.82$ & $59.52 \pm 2.34$ & $78.32 \pm 1.11$  & \underline{$86.93 \pm 0.73$} & \underline{$84.62 \pm 0.98$} & $87.45 \pm 0.36$    & $73.54 \pm 0.87$   \\
        \textbf{$\pi$-GNN}    & $74.94 \pm 1.32$ & \underline{$79.49 \pm 0.56$} & $63.82 \pm 2.19$  & $79.62 \pm 4.52$ & $75.53 \pm 0.65$ & $82.86 \pm 1.66$    & $63.06 \pm 0.60$   \\
        \textbf{GIB}          & $73.60 \pm 2.17$ & $74.73 \pm 2.22$ & $83.80 \pm 1.68$  & $82.41 \pm 2.83$ & $79.52 \pm 3.24$ & $84.94 \pm 1.22$    & $78.29 \pm 0.87$   \\
        \textbf{GSAT}         & $80.45 \pm 2.51$ & $74.38 \pm 0.53$ & $57.72 \pm 1.50$  & $73.68 \pm 3.73$ & $74.95 \pm 2.74$ & \underline{$89.86 \pm 3.02$}    & $57.83 \pm 1.64$   \\
        \textbf{CAL}          & $78.42 \pm 1.82$ & $73.14 \pm 3.62$ & $62.68 \pm 2.14$  & $74.73 \pm 1.42$ & $83.46 \pm 1.42$ & $83.42 \pm 2.24$    & $\mathbf{80.12 \pm 0.52}$   \\ \midrule
        \textbf{GNNExplainer} & $78.26 \pm 0.19$ & $77.52 \pm 2.13$ & $\mathbf{86.27 \pm 2.06}$  & $79.46 \pm 2.68$ & $73.89 \pm 3.57$ & $86.77 \pm 3.70$    & $69.95 \pm 3.08$   \\
        \textbf{SubgraphX}    & $79.53 \pm 2.61$ & $69.59 \pm 1.31$ & $73.41 \pm 2.37$  & $85.27 \pm 3.31$ & $75.38 \pm 3.68$ & $90.16 \pm 2.98$    & $68.53 \pm 3.55$   \\
        \textbf{XGNN}         & $85.47 \pm 2.92$ & $73.43 \pm 2.81$ & $72.39 \pm 2.43$  & $79.38 \pm 5.52$ & $82.89 \pm 0.69$ & $83.75 \pm 0.51$    & $74.16 \pm 1.06$   \\ \midrule
        \textbf{Ours}         & \underline{$86.41 \pm 2.10$} & $\mathbf{82.59 \pm 2.60}$ & \underline{$85.83 \pm 2.17$}  & $\mathbf{91.16 \pm 1.45}$ & $\mathbf{85.33 \pm 3.58}$ & $\mathbf{93.47 \pm 1.64}$    & \underline{$79.08 \pm 1.99$}   \\ \bottomrule
        \end{tabular}
    }
\end{table}

\begin{table*}[!t]
\Huge
    \renewcommand\arraystretch{1.4}
    \caption{Comparison of different methods in terms of consistency.}
    \label{tab:tabel4}
    \resizebox{\textwidth}{!}{
    \begin{tabular}{@{}ccccccccccc@{}}
\toprule
\textbf{Datasets}       & \textbf{ProtGNN} & \textbf{KerGNN}   & \textbf{$\pi$-GNN} & \textbf{GIB} & \textbf{GSAT}     & \textbf{CAL}                     & \textbf{GNNExplainer} & \textbf{SubgraphX} & \textbf{XGNN}                    & \textbf{Ours}        \\ \midrule
\textbf{GraphCycle}     & 0.636$\pm$0.023      & {\underline{0.679$\pm$0.102}} & 0.473$\pm$0.078        & 0.558$\pm$0.124  & 0.631$\pm$0.172       & {0.647$\pm$0.058} & 0.582$\pm$0.064           & 0.489$\pm$0.031        & {0.573$\pm$0.071} & \textbf{0.893$\pm$0.121} \\
\textbf{GraphFive} & 0.484$\pm$0.011      & 0.592$\pm$0.092       & 0.394$\pm$0.023        & 0.429$\pm$0.063  & {\underline{0.751$\pm$0.125}} & {0.363$\pm$0.029} & 0.713$\pm$0.052           & 0.380$\pm$0.026        & {0.461$\pm$0.078} & \textbf{0.802$\pm$0.133} \\ \bottomrule
\end{tabular}
}
\end{table*}

\begin{table}[!t]\Huge
\renewcommand\arraystretch{1.5}
\caption{Comparison of different methods in terms of silhouette score.}
\label{tab:tabel5}
\resizebox{\linewidth}{!}{
    \begin{tabular}{@{}cccccccc@{}}
    \toprule
    \textbf{Methods} & \textbf{ENZYMES}     & \textbf{D\&D}          & \textbf{PROTEINS}    & \textbf{MUTAG}       & \textbf{COLLAB}      & \textbf{GraphCycle}  & \textbf{GraphFive}   \\ \midrule
    \textbf{ProtGNN} & 0.301±0.014          & 0.178±0.021          & 0.614±0.047          & 0.216±0.023          & 0.348±0.045          & 0.237±0.021          & 0.133±0.030          \\
    \textbf{Ours}    & \textbf{0.580±0.042} & \textbf{0.284±0.033} & \textbf{0.791±0.104} & \textbf{0.298±0.033} & \textbf{0.739±0.092} & \textbf{0.480±0.083} & \textbf{0.341±0.025} \\ \bottomrule
    \end{tabular}
}
  \end{table}

\subsection{Qualitative Analysis}
To qualitatively evaluate the performance, we visualize the obtained interative patterns of our framework.

From class perspective, we present the explanations on the synthetic dataset GraphCycle by visualizing part of the interactive patterns of different classes.
The results is shown in Figure~\ref{fig:fig4}(a). 
We can find that our framework manages to learn patterns that are consistent with the ground-truth of ``Cycle'' and ``Non-Cycle''.
For comparison, we also show the identified explaintions of another methods~(ProtGNN) that can provide class-level explanations, the results are shown in Figure~\ref{fig:fig4}(b). 
It can be observed that the explanations identified by ProtGNN do not exhibit distinctiveness across different classes. The reason may lie in the fact that the GraphCycle dataset does not exhibit distinctive properties in local structures, and the method based on subgraph exploration fail to capture the interactions between local substructures, thus resulting in weaker explanations.
Therefore, we believe that our framework is able to unveil representative global patterns.
More results of the explanation from class perspective will be presented in Appendix~\ref{appendix_3}.

From instance perspective, we identify one or more interaction patterns similar to the input graph in the decision-making process of the model to serve as instance-level explanations. 



\subsection{Efficiency Study}
In this section, we compare the efficiency of our proposed framework with several interpretable baselines.
In Table~\ref{tab:tabel6}, we show the time required to finish training for each interpretable model.
It can be observed that the efficiency of our method is only slightly inferior to KerGNN and $\pi$-GNN.
According to the analysis above, our method outperforms both KerGNN and $\pi$-GNN in terms of both prediction performance and explanation performance.
Therefore, we believe that the slight additional time cost is worthwhile.
\begin{table}[!h]
  \Huge
  \caption{Time consumption of different methods. ``*'' indicates the method requires additional pre-training process which takes nearly 72 hours.}
  \vspace{-0.3cm}
  \label{tab:tabel6}
  \resizebox{\columnwidth}{!}{
    \begin{tabular}{@{}crrrrrr@{}}

      \toprule
      \textbf{Methods}   & \makecell[c]{\textbf{ENZYMES}} & \makecell[c]{\textbf{D\&D}} & \makecell[c]{\textbf{COLLAB}} & \makecell[c]{\textbf{MUTAG}} & \makecell[c]{\textbf{GraphCycle}} & \makecell[c]{\textbf{GraphFive}} \\ \midrule
      \textbf{ProtGNN}   & 9590.86s          & 19864.35s    & 34794.25s        & 8920.72s        & 11781.47s            & 4706.38s                 \\
      \textbf{KerGNN}    & 397.89s           & 1357.90s     & 1874.97s         & 400.94s         & 197.01s              & 418.67s                  \\
      \textbf{${\pi}$-GNN}\textsuperscript{*} & 386.06s           & 956.98s      & 1827.25s         & 458.23s         & 256.33s              & 445.37s                  \\
      \textbf{GIB}       & 704.94s          & 2934.93s            & 4210.99s                 & 2977.94s       & 1088.13s           & 1145.76s                        \\
      \textbf{GSAT}      & 452.90s           & 1176.47s     & 2842.64s         & 817.94s         & 599.36s              & 795.29s                  \\ \midrule
      \textbf{Ours}      & 434.12s           & 1021.77s     & 2012.55s         & 469.98s         & 260.99s              & 455.47s                  \\ \bottomrule
      \end{tabular}}
\end{table}

\subsection{Ablation Studies}
In this section, we perform ablation studies of our framework to explore the impact of different experimental setups on the effectiveness of the framework and explore the role of different modules.
Due to space limitations, we only present a portion of results here. More results will be shown in Appendix~\ref{appendix_6}.
\subsubsection{Influence of the Number of Compression Blocks}
\label{section4.5.1}
\begin{figure}[!b]
  \includegraphics[width=\columnwidth]{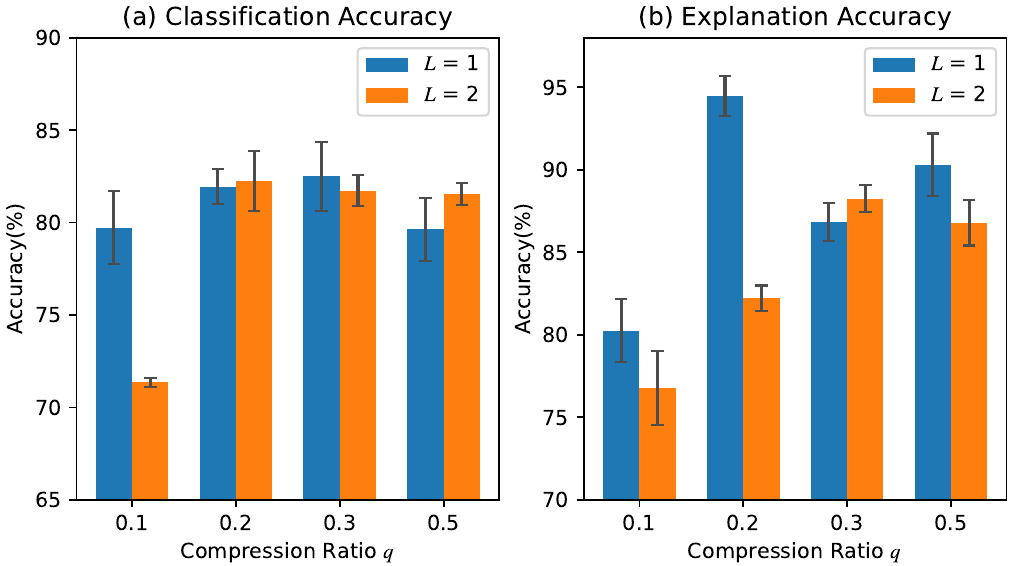}
  \caption{The influence of different number of compression blocks and compression ratio on the model's effectiveness.}
  \Description{The third part of ablation study.}
  \label{fig:fig8}
\end{figure}

First, we investigate the effect of the number of compression blocks $L$ and the compression ratio $q$, where $q$ represents the ratio of the number of nodes after compression to the number of nodes before compression.
We alter the values of $L$ and $q$ as \{1, 2\} and \{0.1, 0.2, 0.3, 0.5\}. 
We conduct experiments on GraphCycle dataset, and the results of classification accuracy and explanation accuracy are presented in Figure~\ref{fig:fig8}.
We can find that when the compression ratio is too high or too low, there is a degradation in both classification accuracy and explanation accuracy.
This may be due to the fact that when the compression ratio is too low, the presence of noisy structures may interfere with the extraction of global information, while a high compression ratio may result in the loss of some information.
Additionally, we also find that the effect of the number of compression blocks on the results varies with different compression ratios.
Therefore, it is crucial to select appropriate number of compression blocks and compression ratios for optimal model performance.

\subsubsection{Influence of the Number of interactive patterns}
\label{section4.5.2}
\begin{figure}[!t]
  \vspace{-0.3cm}
  \includegraphics[width=\columnwidth]{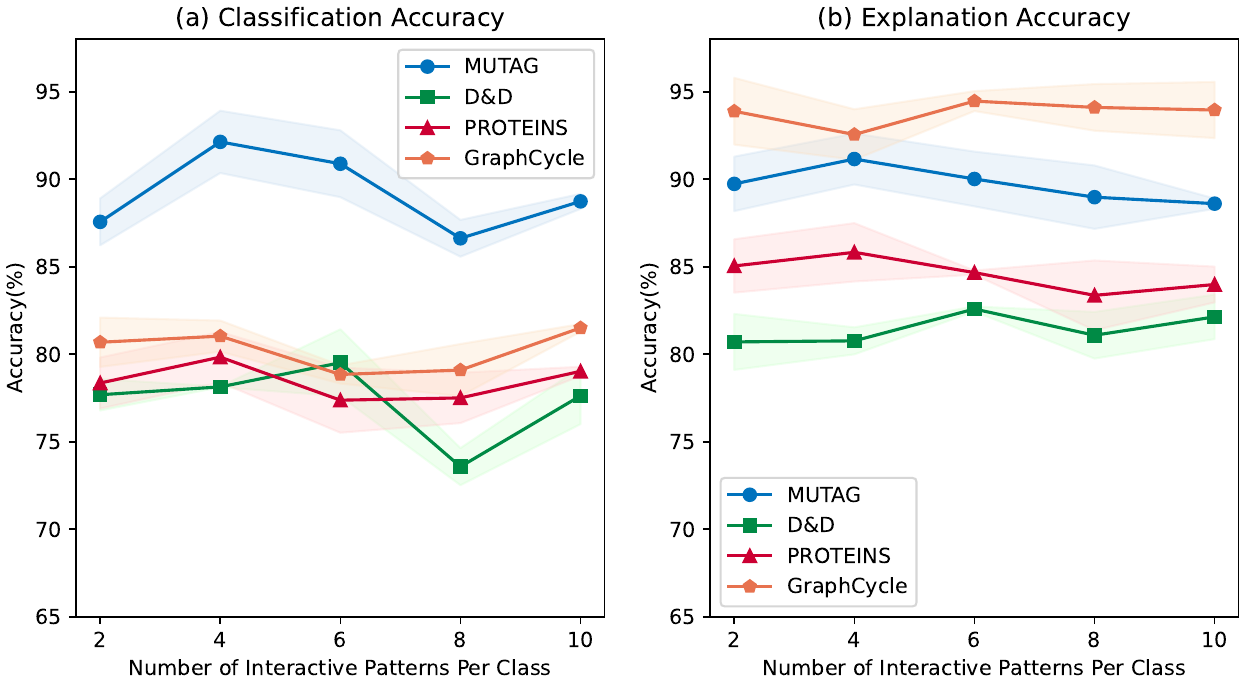}
  \caption{The influence of different numbers of interactive patterns on the model's effectiveness.}
  \label{fig:fig9}
\vspace{-0.3cm}
\end{figure}

Then, we vary the number of interactive patterns per class $T/C$ as \{2, 4, 6, 8, 10\} to investigate its impact to our framework.
We report the results on four datasets in Figure~\ref{fig:fig9}.
We find that with an increase in the number of interactive patterns, both the classification accuracy and explanation accuracy will initially increase and then decrease.
When the number of interactive patterns is too small, they cannot represent all instances in the dataset, resulting in poor prediction performance.
When the number of the interactive patterns is too large, we may obtain excessively similar interactive patterns.
In such cases, the prediction performance may be worse.
The above observations also pave a way for selecting optimal number of interactive patterns in our framework.

\subsubsection{Influence of Different Modules}
\label{section4.5.3}
  \vspace{-0.3cm}
\begin{figure}[!t]
  \includegraphics[width=\columnwidth]{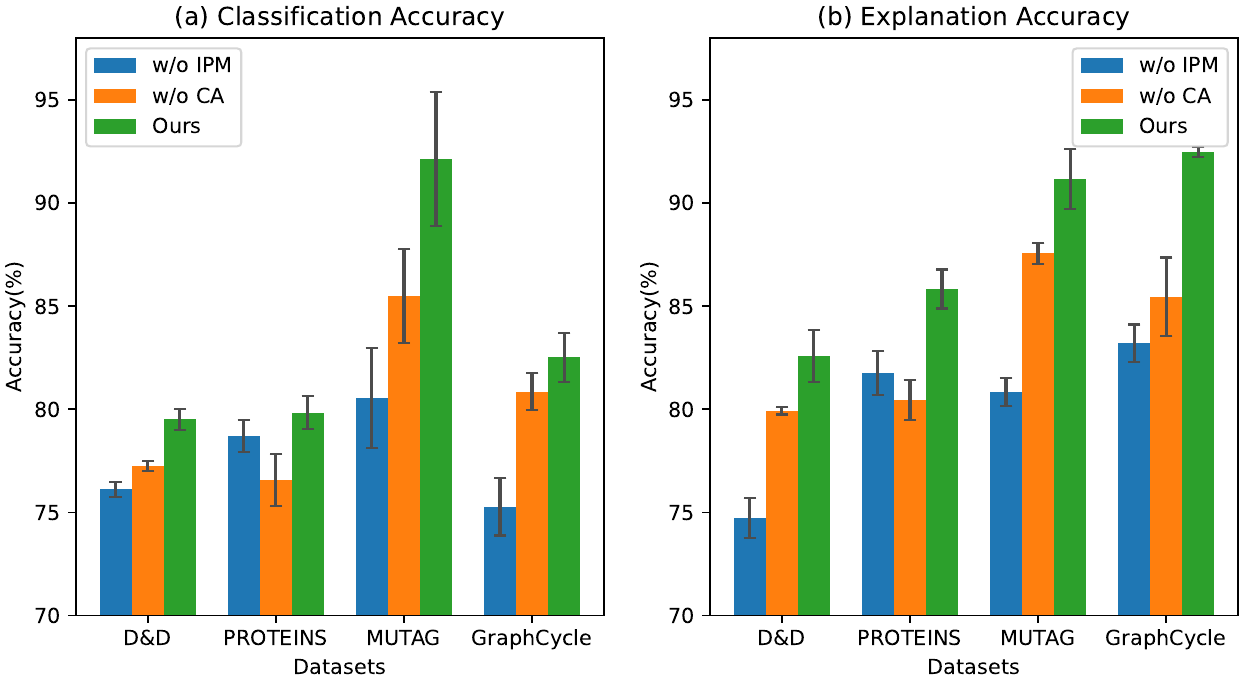}
  \caption{The influence of different modules on the model's effectiveness.}
  \label{fig:fig10}
  \vspace{-0.4cm}
\end{figure}

We adopt clustering assignment module and interactive patterns matching module in our framework.
In order to explore the contribution of these two modules, we implement two variants: (i) without interactive patterns matching module and (ii) without clustering assignment module.
As shown in Figure~\ref{fig:fig10}, we can find that the performance is slightly inferior when the two modules are used individually, while the combination of these two modules achieve the best performance.
Such merits stem from the fact that the combination of these two modules can help to identify the common characteristics in the graphs from the perspective of the global structure interactions, thus effectively enhancing the depth of information mining in graphs.

\section{Conclusion}
In this article, we explore a novel intrinsically explainable graph classification task, called Global Interactive Pattern~(GIP) learning. In contrast to previous methods which focus on exploring local subgraphs for explanation, we propose to analyze cluster-level interaction patterns from a global perspective for attribution analysis.
To this end, we construct a two-stage framework for implementing GIP, by first performing compression of the graph and
then identifying interactive patterns of the coarsened graphs to determine the intrinsic explanations.
Extensive experiments on real-world datasets and synthetic datasets demonstrate the effectiveness of our framework in terms of prediction and explanation performance. This also signifies the value of mining interactive patterns from a global perspective to some extent.
Therefore, our work paves a novel path for interpretable graph classification. In the future, we will further explore this task and endeavor to extend our method to more practical scenarios.



\begin{acks}
This work was supported in part by the Joint Funds of the Zhejiang Provincial Natural Science Foundation of China under Grant LHZSD24F020001, in part by the Zhejiang Province ``LingYan" Research and Development Plan Project under Grant 2024C01114, and in part by the Zhejiang Province High-Level Talents Special Support Program ``Leading Talent of Technological Innovation of Ten-Thousands Talents Program" under Grant 2022R52046.
\end{acks}
\bibliographystyle{ACM-Reference-Format}
\bibliography{sample-base}

\clearpage
\appendix
\setcounter{table}{0}
\setcounter{figure}{0}

\renewcommand{\thetable}{A\arabic{table}}
\renewcommand{\thefigure}{A\arabic{figure}}

\section{More Implementation Details}
\label{appendix_1}
\subsection{Datasets}
\label{appendix_1.1}
\noindent
\textbf{ENZYMES} is a proteins dataset from the BRENDA database~\cite{feragen2013scalable}.
It comes with the task of classifying the enzymes to one out of six EC top-level classes. Specific statistics of the dataset are shown in Table~\ref{tab:tableA1}.

\noindent
\textbf{PROTEINS} is a dataset of proteins from Dobson and Doig dataset~\cite{feragen2013scalable}. It comes with the task of classifying proteins into enzymes and non-enzymes. Specific statistics of the dataset are shown in Table~\ref{tab:tableA1}.

\noindent
\textbf{D\&D}~\cite{dobson2003distinguishing} is a dataset containing high-resolution proteins extracted from a non-redundant subset of the Protein Data Bank. Nodes are amino acids, and two nodes are connected by an edge if the distance between them is less than 6 angstroms. Specific statistics of the dataset are shown in Table~\ref{tab:tableA1}.

\noindent
\textbf{MUTAG}~\cite{wu2018moleculenet} is a molecular property prediction dataset, where nodes are atoms and edges are chemical bonds. Each graph is associated with a binary label based on its mutagenic effect. Specific statistics of the dataset are shown in Table~\ref{tab:tableA1}.

\noindent
\textbf{COLLAB}~\cite{yanardag2015deep} is a scientific collaboration dataset. A graph corresponds to a researcher’s ego network, i.e., the researcher and its collaborators are nodes and an edge indicates collaboration between two researchers. A researcher’s ego network has three possible labels, i.e., High Energy Physics, Condensed Matter Physics, and Astro Physics, which are the fields that the researcher belongs to. Specific statistics of the dataset are shown in Table~\ref{tab:tableA1}.

\noindent
\textbf{GraphCycle} is a self-designed synthetic dataset. Specifically, we first generate 8\textasciitilde15 Barab\'{a}si-Albert graphs as communities, each containing 10\textasciitilde200 nodes. Then, we connect the generated BA graphs in pre-defined two shapes: Cycle and Non-Cycle. To connect nodes in different clusters, we randomly add edges with a probability ranging from 0.05 to 0.15. Specific statistics of the dataset are shown in Table~\ref{tab:tableA1}.

\noindent
\textbf{GraphFive} is a self-designed synthetic dataset.   Specifically, we first generate 8\textasciitilde15 Barab\'{a}si-Albert graphs as communities, each containing 10\textasciitilde200 nodes. Then, we connect the generated BA graphs in pre-defined five shapes: Wheel, Grid, Tree, Ladder, and Star. To connect nodes in different clusters, we randomly add edges with a probability ranging from 0.05 to 0.15. Specific statistics of the dataset are shown in Table~\ref{tab:tableA1}.

\begin{table}[!h]
  \vspace{-0.3cm}
  \caption{The statistics of real-world datasets.}
  \label{tab:tableA1}
  \resizebox{\columnwidth}{!}{
  \begin{tabular}{@{}ccccc@{}}
  \toprule
                    & \textbf{\#Avg.Nodes} & \textbf{\#Avg.Edges} & \textbf{\#Classes} & \textbf{\#Graphs} \\ \midrule
  \textbf{ENZYMES}  & 32.63                & 62.14                & 6                  & 600               \\
  \textbf{D\&D}     & 284.32               & 715.66               & 2                  & 1178              \\
  \textbf{PROTEINS} & 39.06                & 72.82                & 2                  & 1113              \\
  \textbf{MUTAG}    & 17.93                & 19.79                & 2                  & 188              \\
  \textbf{COLLAB}   & 74.49                & 2457.78              & 3                  & 5000              \\ 
  \textbf{GraphCycle}    & 297.70                & 697.18                & 2                  & 2000              \\
  \textbf{GraphFive}   & 375.98                & 1561.77              & 5                  & 5000              \\ 
  \bottomrule
  \end{tabular}}
\end{table}

\subsection{Hyper-parameter Settings}
\label{appendix_1.2}
The hyper-parameters used in our framework include batch size, optimizer, learning rate, epoch, the $\alpha_{1}$ and $\alpha_{2}$ for controlling loss terms in clustering assignment module, the $\alpha_{3}$ and $\alpha_{4}$ for controlling loss terms in interactive patterns matching module, the $\beta_1$ and $\beta_2$ for controlling the contribution of the two modules, etc. The specific settings are presented in Table~\ref{tab:tableA2}.

\begin{table}[!h]\Huge
\renewcommand\arraystretch{1.4}
  \caption{The statistics of hyper-parameters setting.}
  \label{tab:tableA2}
  \resizebox{\linewidth}{!}{
\begin{tabular}{@{}cccccccc@{}}
\toprule
                            & ENZYMES & PROTEINS & D\&D    & MUTAG   & COLLAB  & GraphCycle & GraphFive \\ \midrule
Batch Size                  & 64      & 64       & 128     & 64      & 64      & 128        & 128       \\
Optimizer                   & Adam    & Adam     & Adam    & Adam    & Adam    & Adam       & Adam      \\
Learning Rate               & 0.001   & 0.003    & 0.001   & 0.001   & 0.003   & 0.01       & 0.01      \\
Epoch                       & 500     & 500      & 500     & 500     & 500     & 500        & 500       \\
$\alpha_{1} / \alpha_{2}$   & 0.3/0.2 & 0.2/0.4  & 0.4/0.3 & 0.2/0.4 & 0.3/0.2 & 0.1/0.4    & 0.1/0.1   \\
$\alpha_{3} / \alpha_{4}$     & 0.2/0.2 & 0.4/0.1  & 0.4/0.2 & 0.3/0.1 & 0.1/0.2 & 0.3/0.1    & 0.1/0.1   \\ 
$\beta_{1} / \beta_{2}$     & 0.5/0.4 & 0.5/0.3  & 0.3/0.4 & 0.4/0.5 & 0.4/0.4 & 0.5/0.3    & 0.3/0.4   \\ 
\bottomrule
\end{tabular}}
\end{table}

\section{More Class-level Explanations}
\label{appendix_3}
\begin{figure}[h]
  \vspace{-0.1cm}
  \includegraphics[width=\columnwidth]{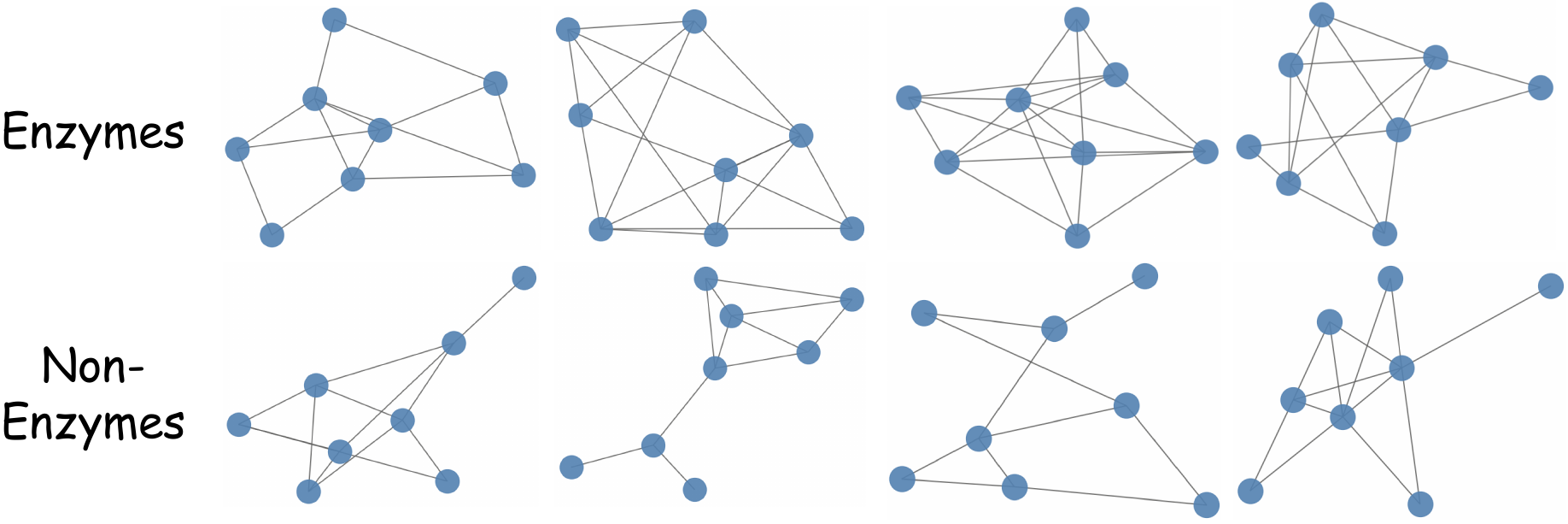}
  \caption{The identified interactive patterns of PROTEINS.}
  \label{fig:figA7}
\end{figure}

\begin{figure}[h]
  \vspace{-0.2cm}
  \includegraphics[width=\columnwidth]{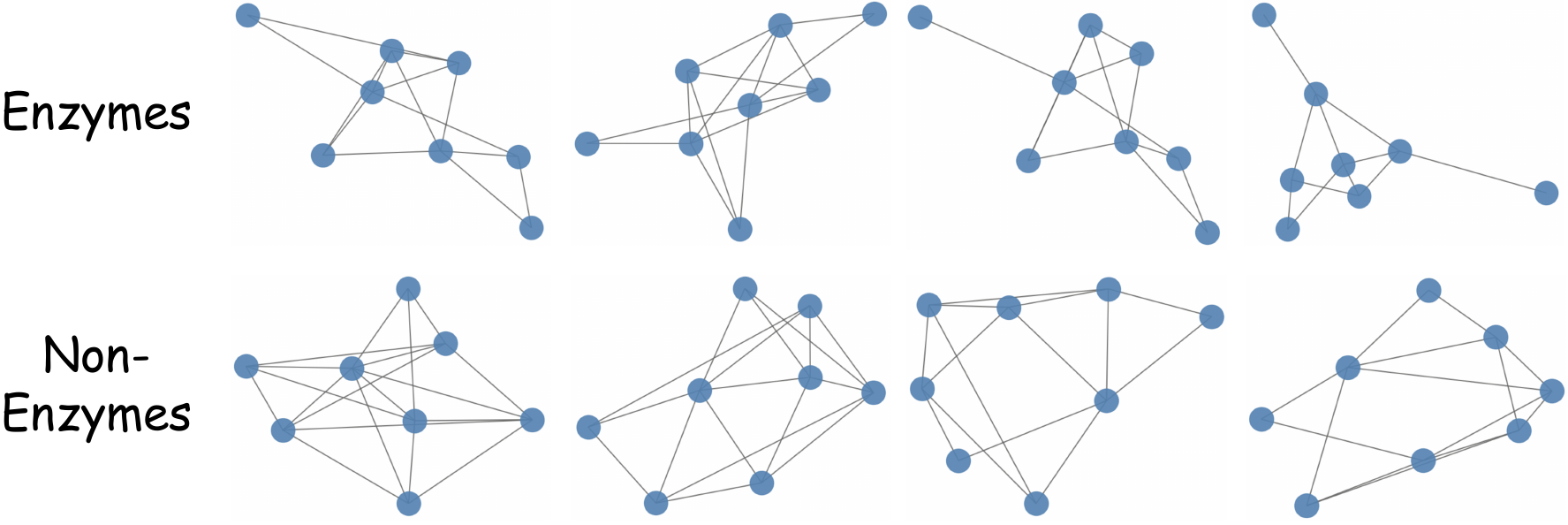}
  \caption{The identified interactive patterns of D\&D.}
  \label{fig:figA8}
\end{figure}

\begin{figure}[!ht]
  \includegraphics[width=\columnwidth]{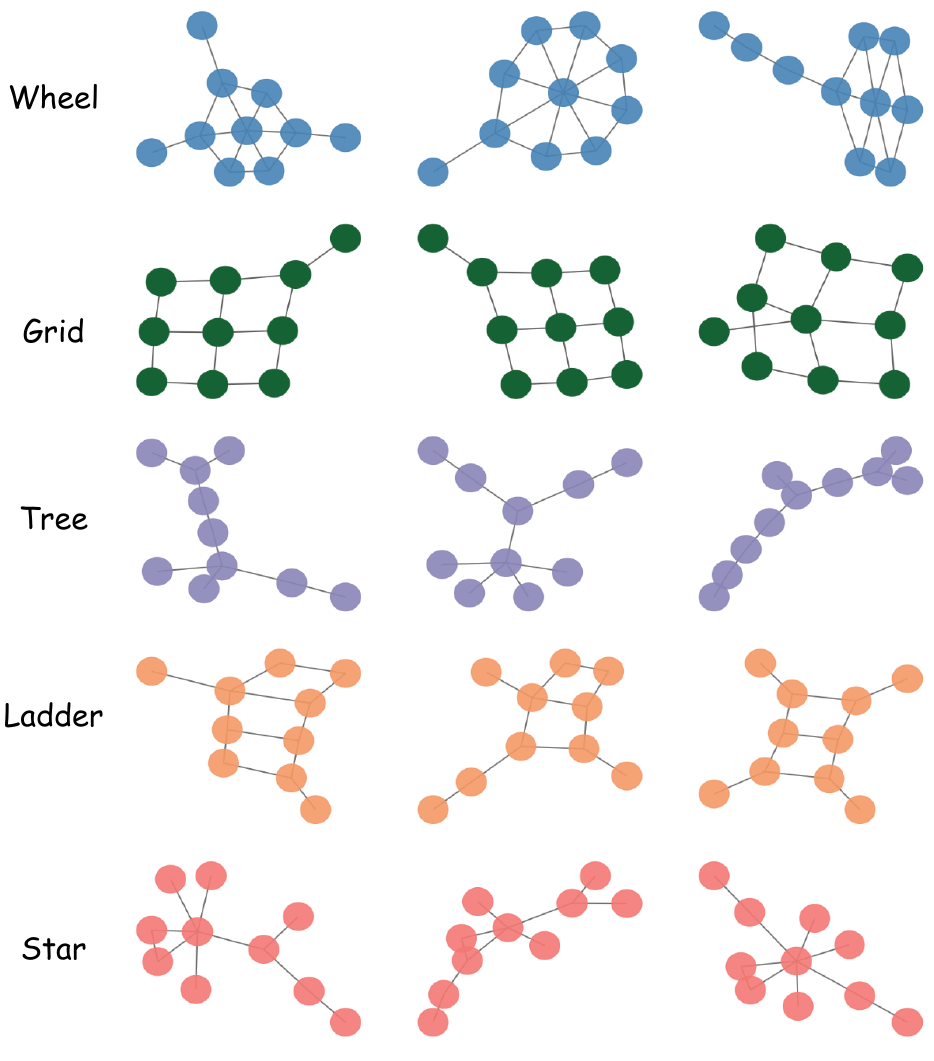}
  \caption{The identified interactive patterns of GraphFive.}
  \label{fig:figA9}
  \vspace{-0.4cm}
\end{figure}

In this section, We will provide more visualization results of class-level explanations on different datasets.
We visualize the global interactive patterns identified in the PROTEINS, D\&D, and GraphFive datasets as explanations from class perspective. The results are shown in Figure~\ref{fig:figA7}, Figure~\ref{fig:figA8}, and Figure~\ref{fig:figA9}.
It can be easily observed that the interaction patterns exhibit commonalities within the same class, while also displaying a certain degree of differentiation between different classes.
For example, in the PROTEINS dataset, the interaction patterns in enzymes exhibit more numerous and longer loops, as well as tighter connections, compared to the interaction patterns in non-enzyme. This observation provides us with new insights to distinguish graphs with different property in the absence of expertise. In the future, we will cooperate with domain experts to conduct more comprehensive analysis.
Similarly, in the GraphFive dataset, the identified interaction patterns in different classes exhibit shapes similar to our pre-defined ground-truth. Therefore, our framework is capable of mining representative interaction patterns in graphs of different classes.

\section{More Ablation Studies}
\label{appendix_6}
\subsection{Influence of the Compression Blocks}
\begin{figure}[!ht]
  \vspace{-0.2cm}
  \includegraphics[width=\columnwidth]{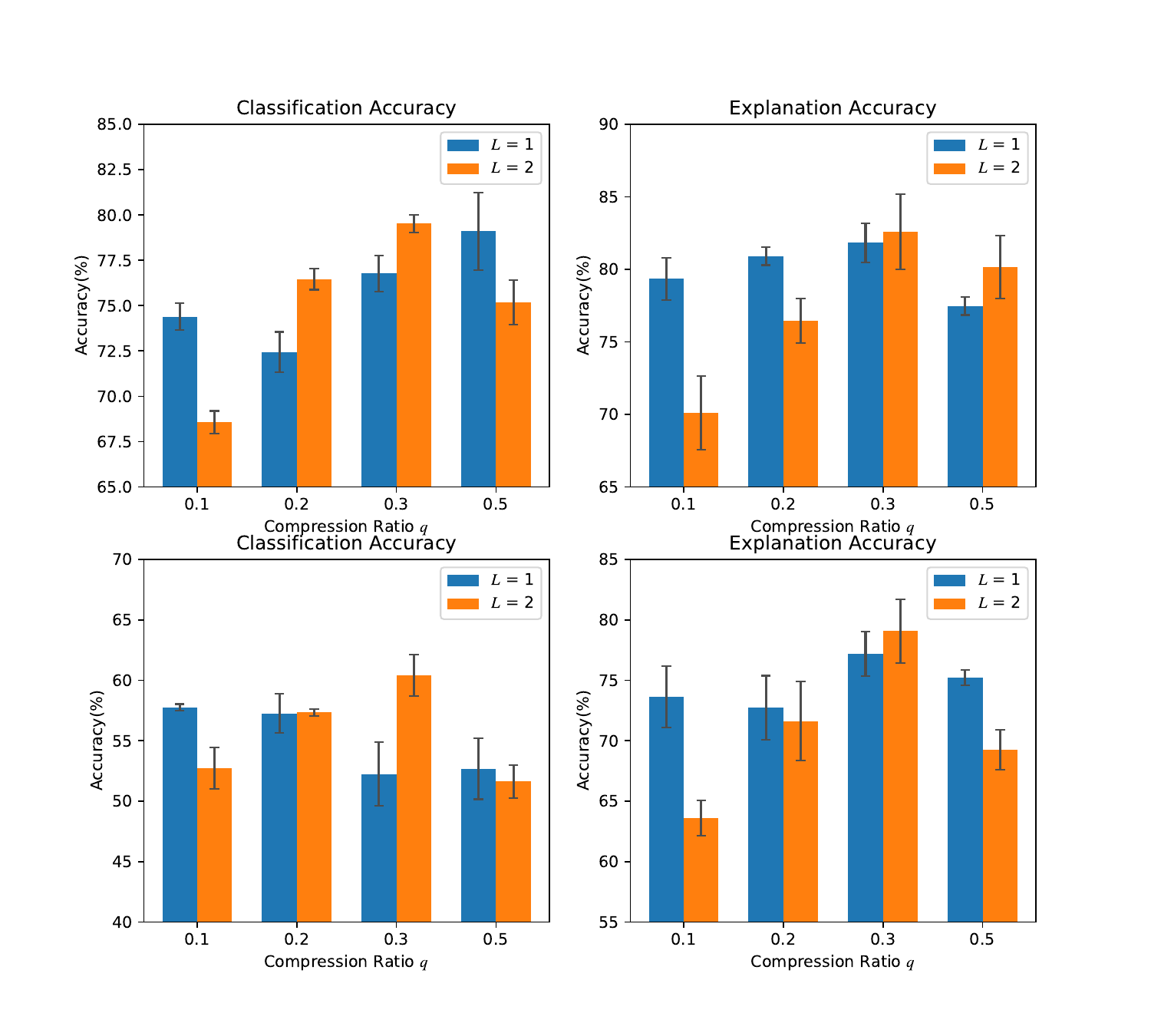}
  \caption{The influence of different numbers of compression blocks and compression ratios on the effectiveness of the model.}
  \label{fig:appendix_ablation1}
  \vspace{-0.2cm}
\end{figure}

In this section, we continue the discussion in Section~\ref{section4.5.1}, and analyze the influence of the number of compression blocks and compression ratios on the model performance with D\&D and GraphFive datasets. We present the results in Figure~\ref{fig:appendix_ablation1}.
It can be seen that for different datasets, the appropriate number of compression layers and compression ratios vary, further confirming the discussion in Section~\ref{section4.5.1}. However, in most cases, fewer compression layers and moderate compression ratios will yield better results.

\subsection{Influence of the Number of Interactive Patterns}
\begin{figure}[!ht]
  \vspace{-0.2cm}
  \includegraphics[width=\columnwidth]{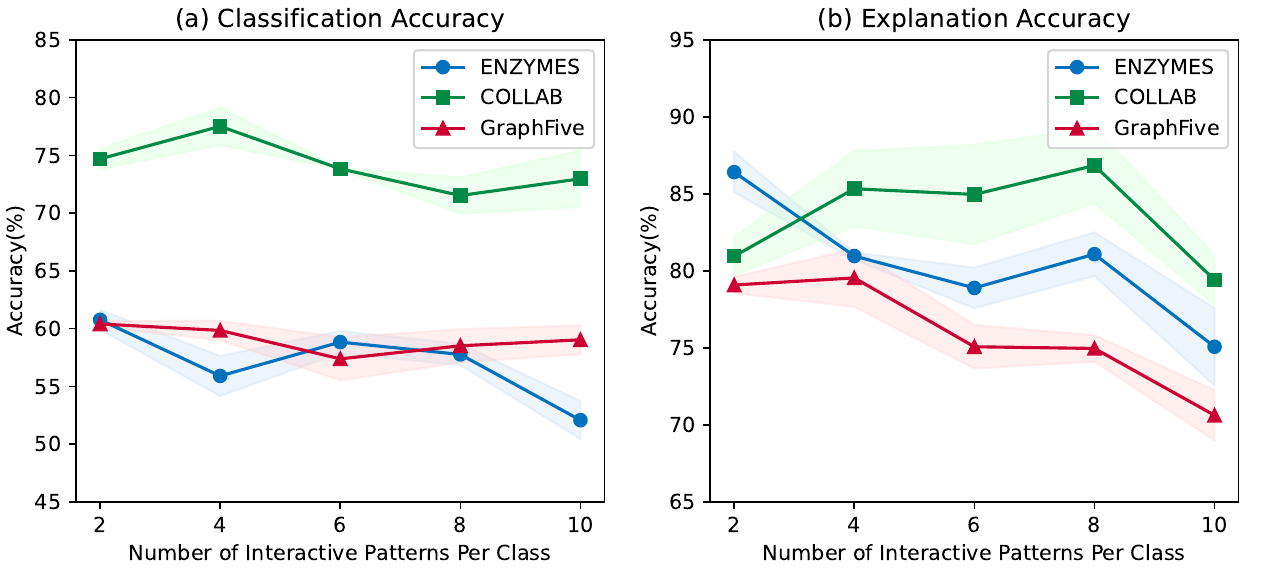}
  \caption{The influence of different numbers of interactive patterns on the model’s effectiveness. The experiments are conducted on ENZYMES, COLLAB and GraphFive.}
  \label{fig:appendix_ablation2}
\vspace{-0.2cm}
\end{figure}

In this section, we supplement the work in Section~\ref{section4.5.2} and demonstrate the variations in model performance with changes in the number of interactive patterns per class on the ENZYMES, COLLAB, and GraphFive datasets. The results are shown in Figure~\ref{fig:appendix_ablation2}.

We further note that changes in the number of interaction patterns have different effects on prediction performance and explanation performance, which requires us to further consider the balance between prediction performance and explanation performance to determine the appropriate number of interaction patterns.

\subsection{Influence of Different Modules}
\begin{figure}[!ht]
  \vspace{-0.2cm}
\includegraphics[width=\columnwidth]{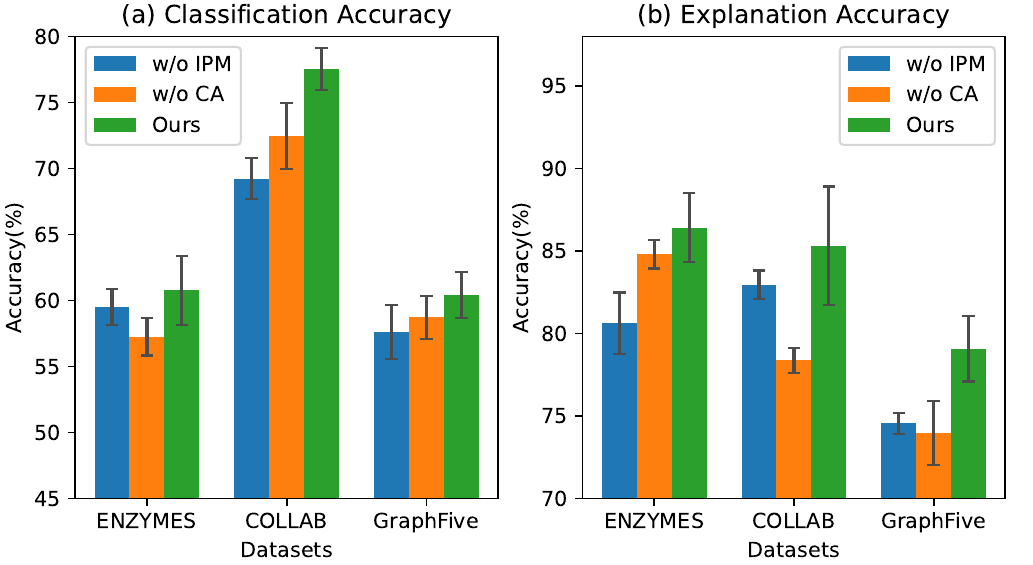}
  \caption{The influence of different modules on the model’s effectiveness.}
  \label{fig:appendix_ablation3}
  \vspace{-0.3cm}
\end{figure}

In this section, we present more results about the influence of different modules on the model performance. The results on ENZYMES, COLLAB, and GraphFive datasets are shown in Figure~\ref{fig:appendix_ablation3}.
These results show the same trend as in Section~\ref{section4.5.3}, i.e., the combination of the two modules achieves better results, which can indicate that our two-stage framework is effective.

\end{document}